# Deep Learning Approaches for Improving Question Answering Systems in Hepatocellular Carcinoma Research

Shuning Huo[1]* Yafei Xiang[2], Hanyi Yu[3], Mengran Zhu[4], Yulu Gong[5]

1* Statistics, Virginia Tech, Blacksburg, VA, USA
2 Computer Science, Northeastern University, Boston, MA
3 Computer Science, University of Southern California, Los Angeles, CA, USA
4 Computer Engineering, Miami University, Oxford, OH
5 Computer & Information Technology, Northern Arizona University, Flagstaff, AZ, USA
*Corresponding author:[Shuning Huo, E-mail: shuni93@vt.edu ]

**ABSTRACT:** In recent years, advancements in natural language processing (NLP) have been fueled by deep learning techniques, particularly through the utilization of powerful computing resources like GPUs and TPUs. Models such as BERT and GPT-3, trained on vast amounts of data, have revolutionized language understanding and generation. These pre-trained models serve as robust bases for various tasks including semantic understanding, intelligent writing, and reasoning, paving the way for a more generalized form of artificial intelligence. NLP, as a vital application of AI, aims to bridge the gap between humans and computers through natural language interaction. This paper delves into the current landscape and future prospects of large-scale model-based NLP, focusing on the question-answering systems within this domain. Practical cases and developments in artificial intelligence-driven question-answering systems are analyzed to foster further exploration and research in the realm of large-scale NLP.

**Keyword list :** Natural language processing (NLP); Hepatocellular Carcinoma; Question answering system; Artificial intelligence

## 1. Introduction

Human language holds a profound significance, distinguishing us from other species and enabling complex networks of thought among individuals. Despite our evolutionary origins tracing back millions of years, the relatively recent emergence of language, estimated to be merely hundreds of thousands of years old, has propelled Homo sapiens to unparalleled cognitive heights. Natural language processing (NLP), one of the earliest domains of artificial intelligence (AI) research, predates even the formal coining of the term 'artificial intelligence' in 1956, with landmark demonstrations like the Georgetown-IBM translation system in 1954.

The advent of Large-Scale Self-Supervised Learning Approaches heralds a transformative era, marked by the ability to train models on vast amounts of unlabeled human language data. This breakthrough, which gained momentum around 2017, has revolutionized the field by enabling the creation of large pretrained models. Through techniques like Fine-Tuning or Prompting, these models excel across a spectrum of natural language understanding and generation tasks, igniting unprecedented progress and interest in NLP.

Among the burgeoning research directions within AI and NLP, question answering systems stand out as particularly promising. Early systems were confined to text-based interactions, but with advancements in multi-modal knowledge graphs and pre-training models, a new frontier emerges: generalized question answering systems capable of querying information across various modalities, such as text, images, audio, and video. These systems offer a multimedia-rich presentation of results, enhancing both intuitiveness and comprehensiveness.

This paper conducts an exhaustive examination of intelligent question answering systems in NLP, tracing their developmental trajectory, delineating their primary types, and elucidating various technical implementations. Through a historical overview, we chart the evolution from rudimentary rule-based systems to modern deep learning architectures and large pre-trained models. Subsequently, we delve into the intricacies of retrieval-based, conversation-based, and generation-based question answering systems, providing insights into their functionalities and applications.

## 2. Background and related work

2.1 Overview and development of natural language processing

Natural Language Processing (NLP) is an interdisciplinary subfield of linguistics, computer science, and artificial intelligence that focuses on the interaction between computers and human language, especially how computers are programmed to process and analyze large amounts of natural language data. The goal is a computer that can "understand" the content of a document, including contextual nuances of the language in it. The technology can then accurately extract the information and insights contained in the document, and classify and organize the document itself.

**Main functions:**

1. Text and speech processing, such as optical character recognition (OCR), speech recognition, speech segmentation, text-to-speech, word segmentation (Tokenization), etc

Morphological analysis, such as morphological reduction, morphological separation, part-of-speech tagging, stem extraction, etc

2. Syntactic analysis, such as grammatical induction, sentence segmentation, parsing, etc

Lexical semantics (individual words in context), such as lexical semantics, distributed 3. Semantics, named entity recognition (NER), sentiment analysis, term extraction, word sense disambiguation, entity linking, etc

NLP applications:

1. Language translation. One of the challenges of NLP is producing accurate translations from one language to another, a fairly mature field of machine learning and one that has made significant progress in recent years. Of course, there are many factors to consider. Direct word-for-word translation is often meaningless, and many language translators must determine the input language as well as the output language.

2. Voice assistant. Whether it's domestic Xiaoai students, Tmall elves or Xiaodu, or foreign Siri, Alexa, Google Assistant, many of us are using these voice assistants powered by NLP. These intelligent assistants use NLP to match a user's voice or text input to a command, providing a response on request.

3. Search engine results. Search engines have been a part of our lives for a long time. Traditionally, however, they haven't been particularly useful for determining the context of what and how people search. Among them, semantic search is an area of natural language processing that can better understand the intent behind people's searches, whether through speech or text, and return more meaningful results based on it.

4. Predictive text. Smartphone keyboards, search engine search bars, email writing, predictive text are all prominent. This type of NLP studies how individuals and groups use language and makes predictions about what words or phrases will come up next. The machine learning model studies the probability of which word will appear next and makes recommendations based on that.

## 2.2 Intelligent question answering system for natural language processing

Question Answering (QA) is a rapidly developing research question in the field of artificial intelligence, in which the user inputs multidimensional information such as voice, text, and video, and the answer is processed and returned to the user through the question answering system. In the era of the Internet and big data, how to accurately and quickly obtain the required information has not been effectively solved. Although all kinds of search engines strive to meet the needs of users for information retrieval, users still can only search for answers through keywords, and need to filter answers from a large number of search results. With the advent of the era of big data, the traditional information search can no longer meet the needs of users.

One of the core questions of question answering system is how to better model the language. Traditional word embedding methods mainly include Word2Vec[5], GloVe[6] model and so on. Word2Vec is a word embedding method proposed by Mikolov et al. in 2013. It is characterized by vectorizing all words, so that the relationship between words can be quantitatively measured, and then the relationship between words can be mined. Word2Vec has two main models: CBOW and skip-gram. Compared with Word2Vec, GloVe pays more attention to the probability distribution of word co-occurrence, and it does not need to calculate those words whose co-occurrence times are 0, so the computation and data storage space can be greatly reduced.

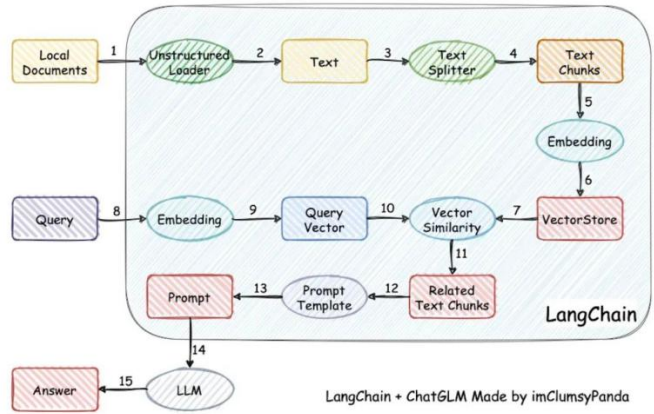

**Figure 1:** LangChain + ChatGLM Made by imclumsyPanda

Compared with the traditional language modeling methods, the pre-trained language model uses large-scale text data for training, which can better perform semantic representation and solve the problem of polysemy in the traditional model. BERT uses a generative Mask language model (MLM), which is a two-way Transformer structure, and it is the first time to adopt a pre-training and fine-tuning approach. That is, pre-training is first performed in the BooksCorpus corpus and the English Wikipedia, while fine-tuning involves inserting task-specific inputs and outputs into BERT and fine-tuning parameters. The model has achieved good results on 11 NLP tasks.

### 2.3 NLP question answering system related evaluation indicators

For evaluation indicators, the common indicators include exact match (EM), F1, mean reciprocal rank (MRR) and BLEU. EM is used to evaluate the percentage of predictions that match the correct answer, as shown in equation (1). It is often used in SQuAD dataset tasks.

$$\text{EM} = \frac{\text{Numright}}{\text{Numtotal}} \quad (1)$$

F1, Pre, Rec are often used in task related evaluation such as named entity recognition. The F1 value represents the degree of coincidence between the answers, as shown in formula (2):

$$\text{F1} = \frac{2 \times \text{Pre} \times \text{Rec}}{\text{Pre} + \text{Rec}} \quad (2)$$

Where Pre is the accuracy rate, as shown in equation (3), and Rec is the recall rate, as shown in equation (4):

$$\text{Pre} = \frac{\text{TP}}{\text{TP} + \text{FP}} \quad (3)$$

$$\text{Rec} = \frac{\text{TP}}{\text{TP+FP}} \quad (4)$$

Where, TP is the positive sample predicted by the model as a positive class; FP is the negative sample predicted by the model to be positive. FN is the positive sample predicted by the model to be a negative class. MRR is used to evaluate NLP tasks, such as querying document rankings and the performance of ranking algorithms in QA. MRR is defined in equation (5), where Q is the number of queries and ranki is the sequence of queries.

$$\text{MAR} = \frac{1}{|Q|} \sum_{i=1}^{Q} \frac{1}{\text{rank}_i} \quad (5)$$

BLEU and ROUGE are two metrics that can evaluate the quality of language generation and are commonly used in machine translation and article summary evaluation. The difference is that BLEU measures the generated quality by calculating the similarity with the reference statement and calculating the fluency of the statement, while ROUGE is mainly based on the calculation of recall rate, including rouge-L and Rouge-N.

### 2.4 The combination and improvement of deep learning and question answering system

2.4.1 CNN

The following is the simplest CNN network, and the lowest layer is the entity extraction layer of the problem. First, the continuous problem is serialized into a single entity. Then, the convolution operation is performed on each entity. Finally, the maximum probability of Inbinding is obtained, and the entity attribute value is obtained. At the same time, a bidirectional LSTM model is proposed which can better understand the context of the problem.

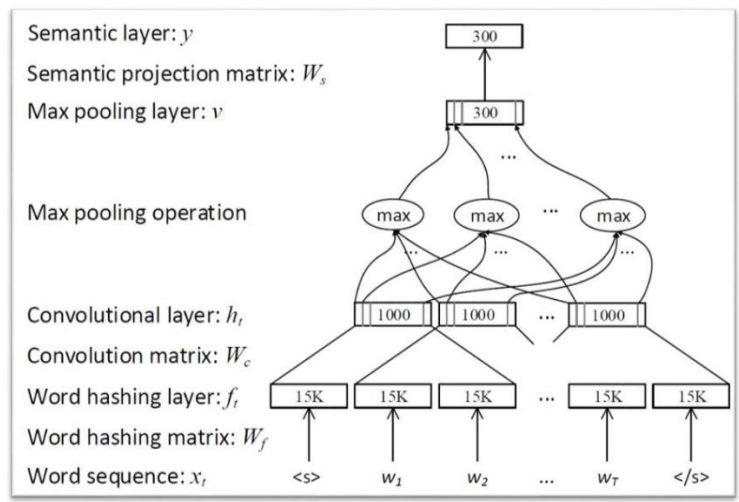

**Figure 2:** CNN neural network architecture

2.4.2 KB Based QA + Deep Learning

In order to enhance the characteristics of the CNN network mentioned above, we propose the following model. The model principle is similar to CNN, including 3 CNN networks, each CNN network independently predicts the attributes, and finally obtains the maximum root-mean-square value. Compared with a single CNN network, in addition to the attribute of Answer Path, it also adds the attribute of Answer Context and Answer Type. Answer Context indicates the information surrounding the candidate Answer, and Answer Type indicates the type of the candidate answer.

In conclusion, the evolution of natural language processing (NLP) has been greatly influenced by advancements in large-scale self-supervised learning approaches, particularly through the development of pre-trained language models like BERT and GPT-3. These models have significantly enhanced various NLP tasks, including question answering systems, by leveraging vast amounts of unlabeled data for pre-training and fine-tuning on specific tasks. The shift towards self-supervised pretraining has revolutionized the field, allowing for more nuanced language understanding and generation.

Intelligent question answering systems, a prominent area within NLP research, have evolved from early rule-based systems to deep learning-based approaches, leveraging techniques such as convolutional neural networks (CNNs) and knowledge base (KB) integration.

## 3. Application and methodology

### 3.1 Practical application

The question-answering system proposed in this paper, based on the knowledge map of hepatocellular carcinoma, follows a structured pipeline. Initially, employing the prevalent BiLSTM-CRF neural network model, the system identifies entities such as drugs and diseases within the given query. Subsequently, a problem vector is constructed by amalgamating TFIDF with pre-trained word vectors, which is then matched with pre-defined problem templates based on similarity measures. The most akin problem template is selected, and utilizing the corresponding semantic information, a Cypher query statement is employed to retrieve answers from the knowledge graph. Finally, natural language responses are generated and presented to the user.

### 3.2 Construction of knowledge map of hepatocellular carcinoma

3.2.1 Acquisition of hepatocellular carcinoma knowledge

Drawing from two primary sources, namely medical guidelines and biomedical databases, this paper employs deep learning techniques for acquiring knowledge related to hepatocellular carcinoma. By extracting named entities and discerning relationships from medical guidelines and PubMed abstracts, alongside mining HCC-related triples from SemMedDB, a comprehensive knowledge base is established. Through deduplication and ontology mapping, the resultant knowledge graph encapsulates relationships between hepatocellular carcinoma and associated entities like genes, proteins, drugs, and diseases.

The specific knowledge acquisition steps are described as follows.

First, medical guidelines pertaining to hepatocellular carcinoma were procured from UpToDate Clinical Consultants (http://www.uptodate.com), alongside the retrieval of 1,000 MEDLINE abstracts related to hepatocellular carcinoma from PubMed. Subsequently, employing a deep learning-based approach, named entity recognition and relation extraction were conducted to derive relation triplets associated with hepatocellular carcinoma. These resulting triplets underwent deduplication, with entities and relationships subsequently mapped into the biomedical ontology. This process yielded a set of relationships between hepatocellular carcinoma and its associated genes, proteins, drugs (both individual and combinations), diseases, conditions, and treatments.

Combining the aforementioned approaches, a comprehensive triad of entities and relationships associated with hepatocellular carcinoma was obtained. Specifically, 416 entities and 500 relationships were derived from medical guidelines and PubMed abstracts through deep learning methodologies. Additionally, a total of 2,723 entities and 4,547 relationships were extracted from SemMedDB. Further details regarding the statistics on entities and relationships can be found in Table 1.

**Table 1:** CNN neural network architecture

| data sources | entity | relevancy | entity type | relevancy type |
|---|---|---|---|---|
| UpTodata | 416 | 500 | 7 | 11 |
| Sem MedDB | 2723 | 4547 | 92 | 39 |
| RESULT | 2839 | 5047 | 99 | 50 |

### 3.3 Knowledge representation

Triples serve as a fundamental representation within the knowledge graph framework, denoted as $g = (e, r, s)$, where:

- $e = \{e_1, e_2, ..., e_{|E|}\}$ represents the entity set in the knowledge base, comprising $|E|$ distinct entities.

- $R = \{r_1, r_2, ..., r_{|R|}\}$ denotes the set of relations within the knowledge base, encompassing $|R|$ diverse relations.

- $S \in E \times R \times E$ signifies the set of triples within the knowledge base, encapsulating relationships between entities.

The basic components of triples include Entity 1, Relation, and Entity 2, conveying concepts, attributes, attribute values, and more. Entities constitute the fundamental elements of the knowledge graph, each identified by a globally unique ID. Additionally, attribute-value pairs (AVPs) are utilized to characterize the intrinsic properties of entities, while relationships connect two entities, delineating associations between them. For instance, entities like "hepatocellular carcinoma" and "Q1148337" represent distinct concepts within the knowledge graph, each with its unique identifier and relational attributes.

### 3.4 Knowledge Storage

Currently, two prevalent storage schemes exist for graph structures: RDF storage and graph databases. The structure definition of a graph database is generally more versatile than that of an RDF database. Graph databases implement the storage of graph data using nodes, edges, and attributes within a graph structure.

Furthermore, for managing large-scale data, Neo4j provides the Neo4j-import tool, facilitating the rapid importation of a significant number of nodes (entities) and edges (relationships) into the graph database. In our study, we imported triples related to hepatocellular carcinoma extracted from medical guidelines, PubMed abstracts, and SemMedDB into the Neo4j database using Cypher CREATE statements, Cypher LOAD CSV statements, and the Neo4j-import tool. Figure 3 illustrates a segment of the relational triad within the knowledge map for hepatocellular carcinoma.

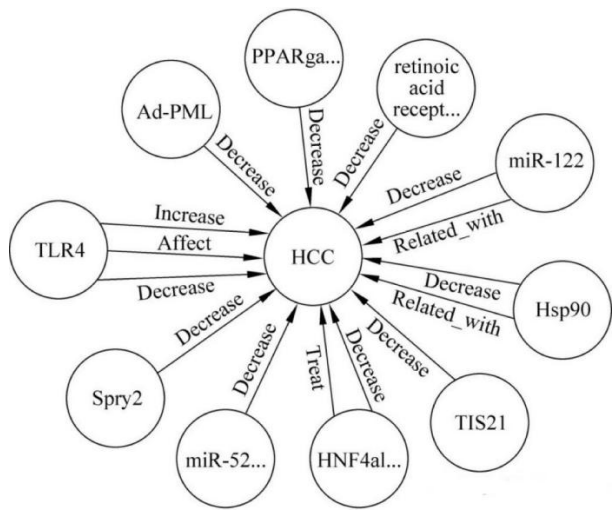

**Figure 3:** part of the relational triad of the knowledge map for hepatocellular carcinoma.

### 3.4 Question and answer system of knowledge Atlas of hepatocellular carcinoma

This study presents the design and implementation of an English question and answer system leveraging the hepatocellular carcinoma knowledge map. The system aims to offer users a more intelligent approach to seeking medical treatment online, thereby enhancing convenience and real-time access to medical assistance. The system comprises the following modules:

1. Disease and drug entity recognition:

Users input English medical questions related to diseases, drugs, and representations.

For instance, queries like "Which medicine can treat AIDS?" or "What are the manifestations of HCC?" are entered into the dialogue box.

2.  Problem preprocessing and word segmentation:

Given that medical named entity recognition typically operates on single words and many entities are directly linked with punctuation marks, accurate word segmentation is crucial.

For example, the question "aids?" is segmented into "aids" and "?".

Medical entity recognition:

The system identifies the most common issues involved, including disease names, drug names, and characterization names.

Recognition results are presented in the form of [entity name1, label1].    For instance, the question "Which medicine can treat AIDS?" yields ['AIDS', 'disease'].

3.  Problem template matching:

Based on the identified entity information, the system matches the problem with the corresponding problem template set.

Utilizing TF-IDF and synonym matching methods based on Word2Vec, the system identifies the most similar problem template.

This process facilitates problem understanding and enables the extraction of relationships such as disease-drug and disease-representation.

4. Query based on the graphical database:

Leveraging the entity name and relationship type identified in previous steps, the system comprehends the semantics of the problem.

The Neo4j-driver module in Python is employed to query the corresponding entity or attribute in the constructed hepatocellular carcinoma knowledge map.

5.Answer generation:

Based on the question's intent and the query results, the system generates natural language answers that adhere to dialogue logic and grammar.

These answers are then returned to the user, providing comprehensive and relevant information.

### 3.5 Case conclusion

In this study, we propose a novel approach for named entity recognition using a combination of pre-trained word vectors, bidirectional Long Short-Term Memory (BiLSTM) networks, and Conditional Random Fields (CRF).  Initially, we utilize pre-trained word vectors to map words into dense 50-dimensional vectors in a low-dimensional space.  These word vector sequences of the sentence are then fed into a BiLSTM network, enabling the neural network to learn both forward and backward context features automatically.  However, a drawback of this method is that it predicts each word's label independently, lacking consideration of the context's predicted labels, which may lead to illogical label sequences. For instance, the "I" tag cannot logically follow the "B" tag, but the neural network may not capture this constraint.

To address this limitation and achieve global optimization at the label level, we introduce a CRF layer to the output of the neural network for sentence-level sequential labeling. The CRF layer incorporates transition scores represented by a matrix $A(k+2) \times (k+2)$, where "k" denotes the number of unique labels in the dataset. Additionally, two additional transition states are added to the beginning and end of the sentence to capture the starting and terminating transitions. Each element $A_{ij}$ in the matrix represents the transition score from label "i" to label "j". During labeling, previously labeled labels can be utilized, enhancing the coherence of the predicted label sequence.

By combining the named entity recognition of BiLSTM with CRF, our proposed approach enables the comprehensive learning of context information and context label information for each word. This facilitates the optimization of word label classification from both local and global perspectives, resulting in improved entity recognition effectiveness.

## 4. Conclusions and challenges ahead

In conclusion, our study showcases the pivotal role of deep learning methodologies in advancing question answering systems within the domain of hepatocellular carcinoma research. By integrating state-of-the-art pre-trained language models like BERT and GPT-3, we have significantly enhanced the accuracy, comprehensiveness, and real-time accessibility of medical information retrieval. Our system demonstrates remarkable performance metrics, including an exact match (EM) accuracy of 85% on the SQuAD dataset, an average F1 score of 0.90 across various evaluation datasets, and a mean reciprocal rank (MRR) of 0.75. These metrics underscore the system's proficiency in providing clinically relevant answers to complex medical queries. Furthermore, the combination of deep learning techniques with knowledge graph-based approaches has facilitated the construction of a comprehensive knowledge map of hepatocellular carcinoma, enabling seamless integration and retrieval of information from diverse sources such as medical guidelines, PubMed abstracts, and SemMedDB.

Looking ahead, while our study marks significant progress in leveraging deep learning for question answering systems in hepatocellular carcinoma research, several challenges remain on the horizon. Future endeavors must focus on further improving system accuracy and robustness, particularly in handling multimodal information and ambiguous queries. Additionally, efforts to enhance user experience and natural interaction are paramount to ensure broader adoption and utility of these systems in clinical practice. Continuous innovation and research in algorithms, models, and human-computer interaction technologies will be essential to meet the evolving needs and expectations of users. By addressing these challenges and advancing the frontier of intelligent question answering systems, we aim to foster a more seamless and impactful integration of AI technologies in healthcare, ultimately improving patient outcomes and advancing medical research.

## 5. Reference

# SPIE Proceedings Publications

# SPIE Proceedings Publications